\documentclass[conference]{IEEEtran}
\IEEEoverridecommandlockouts
\usepackage{cite}
\usepackage{subfig}
\usepackage{amsmath,amssymb,amsfonts}
\usepackage{algorithmic}
\usepackage{graphicx}
\usepackage{textcomp}
\usepackage{xcolor}
\usepackage{comment}
\newcommand{\etal}{\textit{et al.~}}
\usepackage{balance}
\usepackage{multirow}

\def\BibTeX{{\rm B\kern-.05em{\sc i\kern-.025em b}\kern-.08em
    T\kern-.1667em\lower.7ex\hbox{E}\kern-.125emX}}
\begin{document}

\title{Texture CNN for Thermoelectric Metal Pipe Image Classification\\}

\author{\IEEEauthorblockN{Daniel Vriesman}
\IEEEauthorblockA{\textit{Technische Hochschule Ingolstadt} \\
Ingolstadt, Germany \\
daniel.vriesman@carissma.eu}
\\
\IEEEauthorblockN{Alessandro Zimmer}
\IEEEauthorblockA{\textit{Technische Hochschule Ingolstadt} \\
Ingolstadt, Germany \\
alessandro.zimmer@thi.de}
\and
\IEEEauthorblockN{Alceu S. Britto Jr.}
\IEEEauthorblockA{\textit{Pontifical Catholic University of Paran\'{a}} \\
Curitiba, Brazil \\
alceu@ppgia.pucpr.br}
\\
\IEEEauthorblockN{Alessandro L. Koerich}
\IEEEauthorblockA{\textit{\'{E}cole de Technologie Sup\'{e}rieure} \\
Montr\'{e}al, Canada \\
alessandro.koerich@etsmtl.ca}
}

\maketitle

\begin{abstract}

In this paper, the concept of representation learning based on deep neural networks is applied as an alternative to the use of handcrafted features in a method for automatic visual inspection of corroded thermoelectric metallic pipes. A texture convolutional neural network (TCNN) replaces handcrafted features based on Local Phase Quantization (LPQ) and Haralick descriptors (HD) with the advantage of learning an appropriate textural representation and the decision boundaries into a single optimization process.
Experimental results have shown that it is possible to reach the accuracy of 99.20\% in the task of identifying different levels of corrosion in the internal surface of thermoelectric pipe walls, while using a compact network that requires much less effort in tuning parameters when compared to the handcrafted approach since the TCNN architecture is compact regarding the number of layers and connections. The observed results open up the possibility of using deep neural networks in real-time applications such as the automatic inspection of thermoelectric metal pipes.
\end{abstract}


\section{Introduction}
Nowadays it is a fact that computer vision-based solutions are being applied more and more in the industry. One of the branches of these applications aims to release the human operator’s industrial inspection process, offering robust systems with a high-quality performance of quality control and manufacturing process \cite{b1}. 
According to Malamas \etal \cite{b2}, inspection tasks can be categorized with respect to the features into four different groups: dimensional characteristics, surface characteristics, structural quality, and operational quality.  Regarding the surface characteristics, there are several pieces of research in the field of corrosion detection in metallic surfaces. 
This topic requires the attention of the industries, where preventive measures are taken to observe the life reduction of iron/steel components caused by corrosion that could lead to the failure of the system or reduce its efficiency \cite{b3}. A branch that could be used as an example is the thermoelectric companies. Their operation basically consists of metallic components conducting high-pressurized steam through pipelines to generate power. The severe operational conditions subject these pipes to several types of degradation, such as pitting corrosion, material loss, flow accelerated corrosion (FAC) and corrosion cracking \cite{b4}. Once that corrosion effect changes the inner surface of the pipes, it is possible to apply algorithms to extract the texture features in order to classify and evaluate the surface. Vriesman \etal \cite{b5} presented a dataset acquired from pipes corroded in a laboratory setup, which emulates FAC similar to the thermoelectric operational conditions. Besides the dataset, Vriesman \etal \cite{b5} also showed that it is possible, based on handcrafted features, to classify the severity of the corrosion. Once that the handcrafted extraction requires the evaluation of different texture extractors and the adjustment of the corresponding parameters, the aim of this work is to bring to the same subject, an approach based on automatic texture analysis using convolutional neural networks (CNN) in order to identify and classify different types of corrosion conditions.

In a similar direction, some works describe the use of CNN as an automatic feature extractor. For instance, based on an industrial dataset, Tao \etal \cite{b6} described a novel cascade autoencoder that is capable of locating and classifying different defects on metallic surfaces. Such defects are identified via an autoencoder network that learns the representation of the defect data in its convolutional layers and select the features that represent the surface defects, segmenting the regions with the accuracy of 89.60\%. After the segmentation, the defects are then classified by a compact CNN, reaching the state-of-the-art with the accuracy of 86.82\%. Ren \etal \cite{b7} used the Decaf network pre-trained on the ImageNet dataset, characterizing a transfer learning process. The image is segmented in patches and features are extracted based on the full connected layer (fc6) of the Decaf network. The extracted features are used for training a multinomial logistic regression (MLR) model. Their method was evaluated on three public and one industrial dataset, when compared to handcrafted methods such as the multiresolution local binary patterns (MLBP) and the gray level co-occurrence matrix (GLCM), the method improved the accuracy of the classification task between 0.66\% and 25.50\%.

With this in mind, in this paper, our hypothesis is that the use of automatic extraction of deep features may improve the classification accuracy of our previous method for visual inspection of corroded thermoelectric metal pipes \cite{b5}. To such an aim, deep features extracted using a Texture Convolutional Neural Network (TCNN) \cite{b8,b8a} are used to replace well-known and efficient handcrafted features. The experimental results have confirmed our hypothesis, since the accuracy was improved from 98.71\% to 99.20\% in the task of identifying different levels of corrosion in metallic pipes. The deep textural features were able to better deal with some difficulties presented in the images like illumination variance and the presence of blurred spots. In addition, less effort is required in terms of parameters tuning and computational processing when compared to the original handcrafted features since the TCNN architecture is compact regarding the number of layers and connections. The observed results open the possibility to apply the power of deep neural networks to real-time applications such as the automatic inspection.


This paper is organized as follows. Section II describes the feature extraction method and the dataset of thermoelectric pipe images. Section III presents the CNN architecture. Section IV presents the experimental results, which are discussed in Section V. Finally, in Section VI, we present our conclusions and perspectives of future work.  

\section{Representation Learning and Dataset}
\label{sec:rep}
In this section, we describe the dataset of thermoelectric pipe images used in our experiments, and we discussvthe use of handcrafted textural features versus learning textural representation directly from the images.

\subsection{Image Dataset Description}
\begin{figure}[hptb!]
\centering\includegraphics[width=1\linewidth]{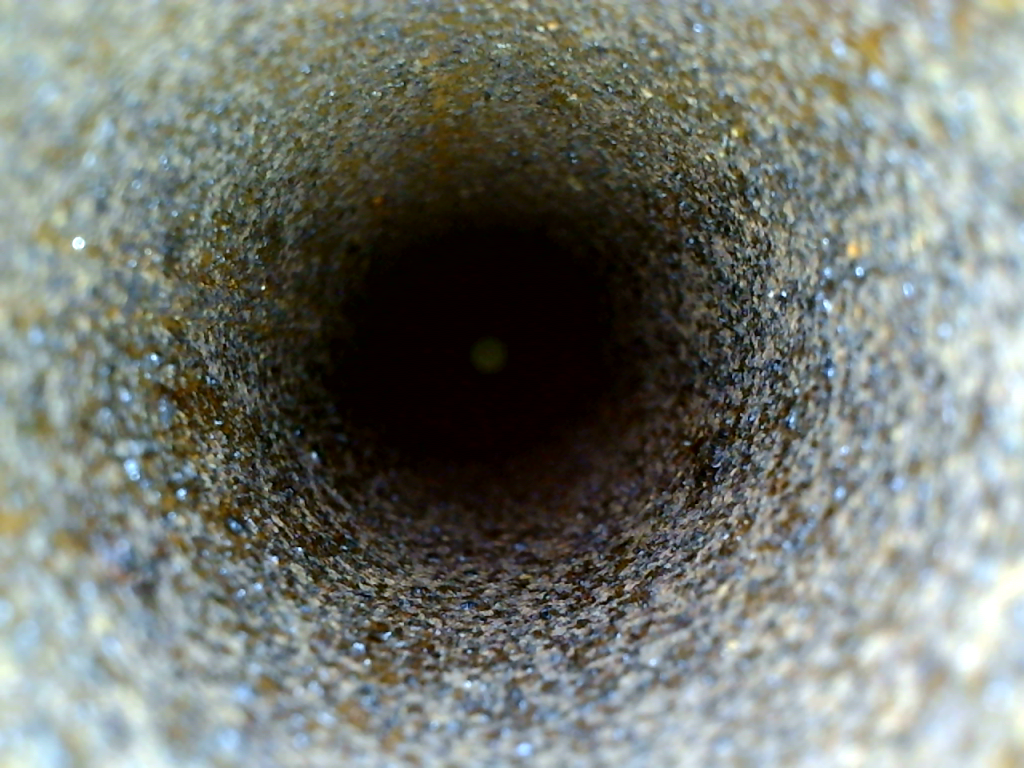}
\caption{An example of a non-preprocessed image of the internal surface of the pipe wall.}
\label{fig:databaseimage}
\end{figure}

\begin{figure}%
    \centerline{
    \subfloat[]{{\includegraphics[height=1.2 \linewidth]{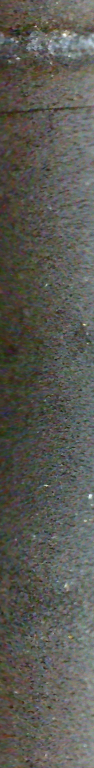}}}
    \qquad
    \subfloat[]{{\includegraphics[height=1.2 \linewidth]{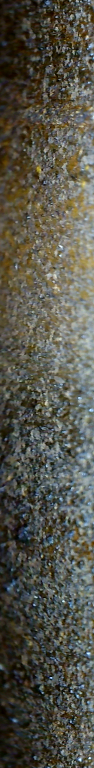}}}
    \qquad
    \subfloat[]{{\includegraphics[height=1.2 \linewidth]{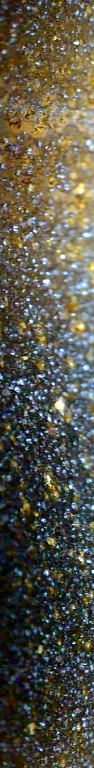}}} }
    \caption{Preprocessed examples representing the three classes of pipes: (a) non-defective (ND); (b) medium corrosion (MC); (c) aggravated corrosion (AC).}%
    \label{fig:preprocessedimages}%
\end{figure}

The dataset of thermoelectric metal pipe images was acquired from the internal surface of pipe walls using the acquisition system described in \cite{b5}. The metal pipes emulate flow accelerated corrosion (FAC) conditions. The images were acquired at the resolution of 1024$\times$768 pixels as shown in Fig.~\ref{fig:databaseimage}. For a better analysis of the focused region, the images were preprocessed using a Cartesian to Log-Polar coordinate conversion to unfold the images. The resulting images have a resolution of 94$\times$768 pixels, as shown in Fig.~\ref{fig:preprocessedimages} \cite{b5}. The dataset consists of three different inner surfaces, which are: non-defective (ND), medium corrosion (MC) and aggravated corrosion (AC), as shown in Fig.~\ref{fig:preprocessedimages}. For each type of inner surface, 50 images were gathered resulting in 150 images.

\begin{figure*}[htbp!]
\centerline{\includegraphics{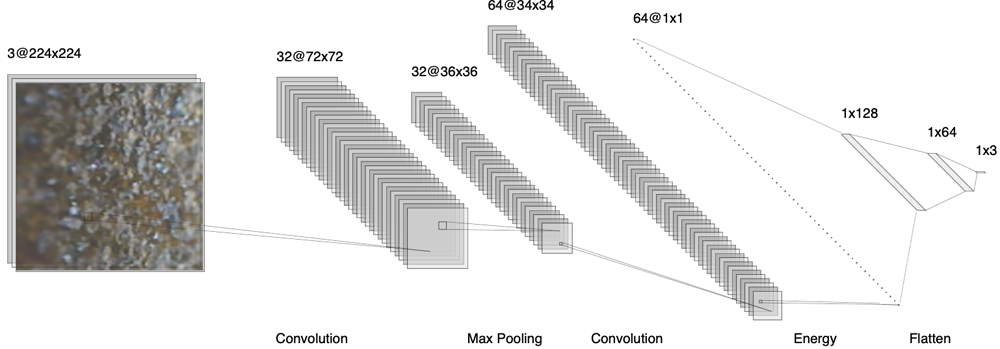}}
\caption{Texture CNN (TCNN) architecture with two convolution layers.}
\label{fig:cnn}
\end{figure*}

A visual inspection of the images has shown that the corrosion changes the piper inner surface. Therefore, the relevant information to discriminate among the three classes is more related to texture than shapes or edges \cite{b5}. The main challenge is that the variability of luminosity and reflectance affect the texture information severely. Besides that, as the acquisition device moves within the pipes while acquiring images, the acquired images could present some blurred spots along with the image. These are some relevant points that a texture extractor must overcome to extract the texture information efficiently in order to train a classifier with a good generalization and robustness.

\subsection{Handcrafted Features vs. Representation Learning}
Several algorithms for extracting textural features from images have been proposed, such as Local Binary Patterns \cite{b9,b9a}, Local Phase Quantization \cite{b10,b10a}, Rotation Pairwise Invariant Co-occurrence Local Binary \cite{b11}, Haralick Descriptors \cite{b12}, and Gabor wavelets \cite{b12b} and other well-known algorithms \cite{b12a}. The point is that in a handcrafted approach, the designed solution should consider the balance between accuracy and computational efficiency. The robustness of these algorithms normally comes with high computational costs. Furthermore, these are general representations that may not 
take into account all statistical properties and repeated patterns since these algorithms have several parameters to tune in order to achieve a good performance.

For example, to overcome the illumination/reflectance variation and the presence of blurred spots using a handcrafted extractor, Vriesman \etal \cite{b5} used as one of the extractors the Local Phase Quantization (LPQ) algorithm, which is robust for blurred spots and illumination invariant, reaching accuracy between 87.43\% and 96.28\% depending on the values of two parameters: slice width and window dimension. Furthermore, to achieve a higher performance, the LPQ features had to be concatenated with Haralick Descriptors (HD), reaching for the best setup the accuracy of 98.71\%. However, the use of two different descriptors requires the fine-tuning of three parameters as well as to find the best concatenation of feature representations. Besides that, the fine-tuning of the feature extractors must also be coordinated with the parameters of the classifier algorithm. Therefore, finding the best setup to maximize performance in terms of accuracy requires many efforts in terms of parameter tuning.  


Convolutional neural networks (CNN) emerged as an interesting alternative since they can learn both the representation and the classification into a single optimization process. Nani \etal \cite{b13} have shown that along with the layers of a CNN, an image can be represented at different levels of abstraction from low to high-level features, which can provide greater robustness to intra-class variability and the presence of blurred spots and illumination variation. In case of blur situations, Wang \etal \cite{b14} described the performance of a Simplified-Fast-AlexNet (SFA) to classify different types of blurred images, such as defocus, Gaussian, and motion blur. They compared the performance between consolidated handcrafted methods and the trained model over the same dataset. The SFA solution reached accuracy between 93.75\% and 96.66\%, while the handcrafted methods reached accuracy below 90.00\%, proving the capability of the CNN approaches to deal with blurred spots. Considering the illumination/reflectance variation, Ramaiah \etal \cite{b15} used a CNN approach that improved the accuracy in 4.96\% over the usual handcrafted methods for facial recognition under non-uniform illumination, proving the capability of the CNN to classify correctly images in an illumination invariant way.

Therefore, in this paper we exploit the capacity of CNNs to learn good representations and good discriminators to deal with the classification of thermoelectric metal pipe images, which contain essentially textural information. 

\section{Texture CNN Architecture}
The architecture of the Texture CNN is based on the T-CNN proposed by Andrearczyk and Whelan \cite{b8} which includes an energy layer that pools the feature maps of the last convolutional layer by calculating the average of its rectified activation output. This results in one single value per feature map, similar to an energy response to a filter bank.

\begin{figure*}[htbp!]
\centerline{\includegraphics[width=0.62\linewidth]{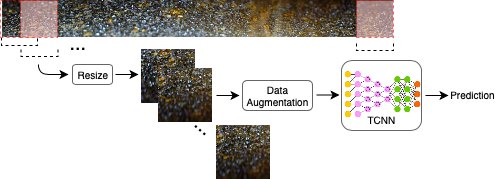}}
\caption{A sliding window with 50\% overlapping to extract squared images of 94$\times$94 which are further upscaled to 224$\times$224 to feed the  TCNN.}
\label{fig:sliced}
\end{figure*}

The proposed Texture CNN uses a simple structure derived from AlexNet network, as illustrated in Fig.~\ref{fig:cnn}, but it has a reduced number of parameters made up of only two convolutional layers, a max pooling layer to reduce the spatial size of the representation, two fully connected layers and an output layer. Besides the reduced number of layers, this CNN achieves a good performance in learning textural features while requiring low computational time and memory size. This trade between performance and computational time is possible due to the energy layer (Fig.~\ref{fig:cnn}). 
The implementation of this layer was made in a way that preserves the data flow of the original layers. That means that the flattened output of the energy layer is redirected directly after the last pooling layer, in the concatenation layer. This concatenation generates a new flattened vector containing information from the shape of the image and the texture, which are then propagated through the full connected layers \cite{b8}. 

The advantage when considering a feature extraction approach based on such a CNN is the fact that each layer learns through the forward and backpropagation procedure, an appropriate representation, which can lead to a better classification of unknown images. This relevant point is where this kind of approach differs from the usual handcrafted methods, where the extraction and selection of features are based on changing the parameters of the algorithms and training classifiers separately.

\subsection{Pre-Trained Models}
From the perspective of deep learning, the classification problem can also be solved through transfer learning. Instead of starting the learning process with randomly initialized networks, we start with a pre-trained model that was trained on a large dataset to solve a problem similar to the one we want to solve. Several pre-trained models used in transfer learning are based on large CNNs that have a large number of parameters. VGG16, Inception V3, and ResNet50, all trained on Imagenet dataset, which are among the pre-trained CNN models used for transfer learning, have 138M, 23M and 25M of parameters respectively.

However, these pre-trained CNNs focus on obtaining information about the overall shape of the image which leads to sparse and complex features that are less appropriate in texture analysis as we mainly seek repeated patterns of lower complexity \cite{b8}. Nevertheless, we have also evaluated some pre-trained models in Section IV.


\subsection{Dataset Preparation}
For training and testing the CNN in the thermoelectric metal pipe dataset, some modifications were necessary to evaluate the proposed method. Each preprocessed image present on the dataset (Fig.~\ref{fig:preprocessedimages}) is upscaled using a bicubic interpolation over 4$\times$4 pixel neighborhood to match the input dimensions required by the input layer of CNNs. Furthermore, to train the CNNs properly, it is necessary large amounts of data. To such an aim, we used the sliding window with overlapping of 50\%. Therefore, a 94$\times$768 image produces fifteen 94$\times$94 images, resulting in 2,250 images. Furthermore, during the training, we generate batches of tensor image data with real-time data augmentation. The low-level data augmentation employs transformations such as horizontal flipping, rotation, and width and height shifting. This process can be observed in the Fig.~\ref{fig:sliced}. 

\section{Experimental Results}
The evaluation of the TCNN was carried out using two different methodologies, 3-fold cross-validation (CV) and hold-out. For the 3-fold CV, one fold was used each time for training, validation and test, resulting into three TCNNs. The percentage of images in each fold, as well as the accuracy when each fold was used for training, validation and test is shown in Table~\ref{tab:kfold}.

\begin{table}[htpb!]
\centering
\small
\begin{tabular}{c c c c c }
\hline
\textbf{Fold} &\textbf{Images} & \multicolumn{3}{c}{ \textbf{Accuracy (\%)}}\\
 \textbf{} & \textbf{(\%)} & \textbf{Training} & \textbf{Validation} & \textbf{Test}\\
\hline
1 & 34 & 99.50 & 94.20 & 94.80 \\
2 & 34 &99.60 &97.90 &98.80 \\
3 & 32 & 100.0 &99.90 &96.80 \\
\hline
\end{tabular}
\caption{Accuracy of the proposed TCNN using 3-fold cross-validation method.}
\label{tab:kfold}
\end{table}

The average accuracy on the test set is 96.80\% with a standard deviation of 2.00\%. Besides the 3-fold CV, the TCNN was also evaluated using the hold-out method. For such an evaluation, we just merged folds 1 and 2 to make our training set (68\% of samples), and use the fold 3 as a test set (32\%). During the training procedure, 20\% of the training set was used as validation set to look at the mean squared error and stop the training to avoid overfitting. This validation set was also used to tune the hyperparameters of the feature extraction algorithms (LPQ and HD) and SVM \cite{b5}. The percentage of images used, and the accuracy of each step can be observed in Table~\ref{tab:holdout}. The accuracy on the test set is 99.20\% using the hold-out method.

\begin{table}[htpb!]
\centering
\small
\begin{tabular}{l c c}
\hline
\textbf{Dataset} & \textbf{Images} & \textbf{Accuracy} \\
                 & \textbf{(\%)}   & \textbf{(\%)} \\
\hline
Training    & 54.4  &    99.60\\
Validation  & 13.6  &    99.40\\
Test        & 32.0  &    99.20 \\
\hline
\end{tabular}
\caption{Accuracy of the proposed TCNN using hold-out method.}
\label{tab:holdout}
\end{table}

Table \ref{tab:cnns} shows the performance achieved by some pre-trained models after fine-tuning them on the thermoelectric metal pipe dataset. The fine-tuning consists of replacing the fully connected layers and the output layer by two fully connected layers (128 and 64 neurons) and an output layer (3 neurons) with softmax activation (similar to the three last layers of the TCNN shown in Fig.~\ref{fig:cnn}). Table \ref{tab:cnns} also shows the size and number of trainable parameters of the fine-tuned models after adapting their last layers to the target problem. The proposed TCNN is the most compact model which has the lowest number of trainable parameters and which leads to the highest accuracy.


\begin{table}[htpb!]
\centering
\small
\begin{tabular}{l r r l}
\hline
\textbf{Model} & \textbf{Size} & \textbf{\# Trainable} &  \textbf{Accuracy} \\
                 & \textbf{(MB)}   & \textbf{Parameters} & \textbf{(\%)}  \\
\hline
\bf TCNN            & \bf 0.6    &    \bf 43,267   & \bf 96.80 $\pm$ 2.00\\ 
VGG16           & 148     &    10,295,042  & 95.66 $\pm$ 2.57\\
ResNet50        & 120     &    3,682,051 & 80.86 $\pm$ 6.25\\
InceptionV3     & 144     &    17,672,835  & 50.82 $\pm$ 11.16\\
MobileNetV2     & 101     &    8,032,515  & 50.71 $\pm$ 10.24\\
\hline
\end{tabular}
\caption{Comparison of some pre-trained models after fine-tuning to the problem of thermoelectric metal pipes and the proposed TCNN using the 3-CV method.}
\label{tab:cnns}
\end{table}

\begin{figure*}[hptb!]
\centering
\begin{tabular}{c}
\includegraphics[width=1\linewidth]{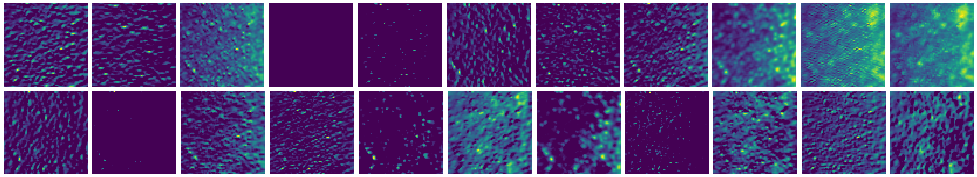} \\
(a)\\
\includegraphics[width=1\linewidth]{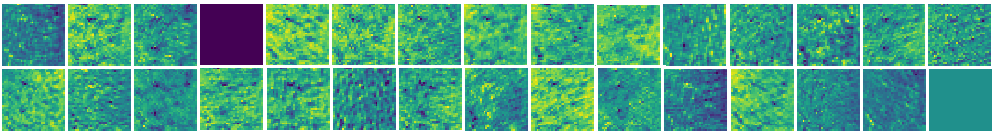} \\
(b)\\
\end{tabular}
\caption{Selected activation maps learned resulting from the convolution operation at the (a) first convolutional layer (72$\times$72); (b) second convolutional layer (34$\times$34).}
\label{fig:convfilters}
\end{figure*}

\section{Discussion}
Table~\ref{tab:3} shows the classification accuracy achieved by the proposed TCNN as well as the results achieved by other approaches based on handcrafted features and an SVM classifier.

\begin{table}[htpb!]
\centering
\small
\begin{tabular}{l  c}
\hline
\textbf{Feature \&  Classifier}  & \textbf{Accuracy (\%)}\\
\hline
Proposed TCNN & \textbf{99.20}\\
VGG16 & 99.10 \\
LPQ$_{9,3}$+HD$_{9}$ and SVM \cite{b5} & 98.71 \\
LPQ$_{9,3}$+HD$_{9}$ and LDA \cite{b5} & 98.18\\
ResNet50 & 84.17 \\
InceptionV3 & 66.39 \\
MobiliNetV2 & 63.47 \\
\hline
\end{tabular}
\caption{Accuracy achieved by the proposed TCNN and other approaches based on pre-trained CNNs and handcrafted features using the hold-out method.}
\label{tab:3}
\end{table}

It is important to highlight that we cannot make a direct comparison between the results achieved by the TCNN with the results achieved in our previous work. Besides the differences in the size and in the shape of the images used to extract the features, the images of each data split are not exactly the same. Having said that, the high accuracy of the CNN approach shows the feasibility of implementing a different solution for the same task based on learning a textural representation directly from data. Concerning the handcrafted features presented in our previous approach based on LPQ and Haralick Descriptors (HD) \cite{b5}, the accuracy of the models varies according to the image size and the parameters of the LPQ algorithm (represented by the subscripted numbers in LPQ and HD). Furthermore, the best accuracy was achieved by fusing the information extracted from both extractors for an image with a determined size, as indicated in Table~\ref{tab:3}. The variance in the accuracy can lead to a parameter-dependent solution, which means the need of time-consuming procedure for tuning the hyperparameters of feature extractors and classifiers according to the changes in the dataset, such as the inclusion of new classes or different image resolutions. On the other hand, the TCNN is able to learn appropriate features and decision boundaries through forward and backward propagation.

The TCNN leads to a better generalization through the learning of the extraction layers, handling with efficiency the illumination variance and the presence of the blurred spots. Figs.~\ref{fig:convfilters}a and \ref{fig:convfilters}b show some activation maps resulting from the first and second convolutional layer, respectively. The activation maps are useful for understanding how the convolution layers transform the input. The first layer is retaining (high energy) directional textures, although there are several filters that are not activated and are left blank. At the second convolutional layer, the activation maps encode higher-level concepts (low-energy) carry increasingly less directional visual contents.

\section{Conclusion}
In this paper, we presented a texture convolutional neural network (TCNN) for automatic visual inspection of thermoelectric metal pipes that learns an appropriate textural representation and the decision boundaries into a single optimization process. The experimental results have shown that the TCNN outperforms a previous approach based on handcrafted features and achieves the accuracy of 99.20\% in the task of identifying different levels of corrosion in the inner surface of thermoelectric pipe walls. Furthermore, the TCNN is a compact network that requires much less effort in tuning parameters when compared to the handcrafted approach since its architecture is compact regarding the number of layers and connections. The observed results open up the possibility of using deep neural networks in real-time applications such as the automatic inspection of thermoelectric metal pipes.

The proposed approach could bring potential scalability for real applications in terms of visual inspections, due to CNN capability of generalization by learning a textural representation efficiently through its learning process. Besides that, usually, when the topic is convolutional neural networks and deep neural networks, it is normal to think in powerful backends with high computational power and GPUs. But regarding the context of industries, a solution could be implemented in compact or embedded systems where there are not many computational resources. Considering this, the paper brings a solution using a compact TCNN, where the optimization of the layers and the addition of a new layer offers high accuracy in the classification task at low computational cost, which is a good aspect for real-time applications and embedded systems \cite{b5}.


\section*{Acknowledgment}
The authors of this work acknowledge the ANEEL for the Research and Development program, the Neonergia Group, for the project funding and the LACTEC for the infra structure and support.






\balance


\end{document}